\documentclass{article}

\usepackage{microtype}
\usepackage{graphicx,amsfonts,amsmath}
\usepackage{subfigure}
\usepackage{breqn}
\usepackage{booktabs} 

\newcommand{\tens}[1]{{\cal{#1}}}
\def\tN{{\tens{N}}}

\usepackage{hyperref}
\hypersetup{colorlinks,allcolors=black}

\usepackage[accepted]{icml2021}

\icmltitlerunning{HyperHyperNetworks for the Design of Antenna Arrays}

\begin{document}

\twocolumn[
\icmltitle{HyperHyperNetworks for the Design of Antenna Arrays}



\icmlsetsymbol{equal}{}

\begin{icmlauthorlist}
\icmlauthor{Shahar Lutati}{equal,to}
\icmlauthor{Lior Wolf}{equal,to,goo}
\end{icmlauthorlist}

\icmlaffiliation{to}{Tel Aviv University}
\icmlaffiliation{goo}{Facebook AI Research}

\icmlcorrespondingauthor{Shahar Lutati}{shahar761@gmail.com}
\icmlcorrespondingauthor{Lior Wolf}{liorwolf@gmail.com}

\icmlkeywords{Machine Learning, ICML}

\vskip 0.3in
]



\printAffiliationsAndNotice{ } 

\begin{abstract}

We present deep learning methods for the design of arrays and single instances of small antennas. Each design instance is conditioned on a target radiation pattern and is required to conform to specific spatial dimensions and to include, as part of its metallic structure, a set of predetermined locations. The solution, in the case of a single antenna, is based on a composite neural network that combines a simulation network, a hypernetwork, and a refinement network. In the design of the antenna array, we add an additional design level and employ a hypernetwork within a hypernetwork. The learning objective is based on measuring the similarity of the obtained radiation pattern to the desired one. Our experiments demonstrate that our approach is able to design novel antennas and antenna arrays that are compliant with the design requirements, considerably better than the baseline methods. We compare the solutions obtained by our method to existing designs and demonstrate a high level of overlap. When designing the antenna array of a cellular phone, the obtained solution displays improved properties over the existing one.
\end{abstract}

\section{Introduction}
\label{sec:intro}

Since electronic devices are getting smaller, the task of designing suitable antennas is becoming increasingly important \cite{anguera_advances_2013}.  
However, the design of small antennas, given a set of structural constraints and the desired radiation pattern, is still an iterative and tedious task \cite{miron_small_2014}.  Moreover, to cope
with an increasing demand for higher data rates in dynamic communication channels, almost all of the current consumer devices include antenna arrays, which adds a dimension of complexity to the design problem \cite{Bogale_massive_mimo}.

Designing antennas is a challenging inverse problem: while mapping the structure of the antenna to its radiation properties is possible (but inefficient) by numerically solving Maxwell's equations, the problem of obtaining a structure that produces the desired radiation pattern, subject to structural constraints, can only be defined as an optimization problem, with a large search space and various trade-offs~\cite{H_small_antenna}. Our novel approach for designing a Printed Circuit Board (PCB) antenna that produces the desired radiation pattern, resides in a 3D bounding box, and includes a predefined set of metallic locations. We then present a method for the design of an antenna array that combines several such antennas. 


The single antenna method first trains a simulation network $h$ that replaces the numerical solver, based on an initial training set obtained using the solver. 
This network is used to rapidly create a larger training set for solving the inverse problem, and, more importantly, to define a loss term that measures the fitting of the obtained radiation pattern to the desired one. The design networks that solve the inverse problem include a hypernetwork~\cite{ha2016hypernetworks} $f$ that is trained to obtain an initial structure,  which is defined by the functional $g$. This structure is then refined by a network $t$ that incorporates the metallic locations and obtains the final design. 
For the design of an antenna array, on top of the parameters of each antenna, it is also necessary to determine the number of antennas and their position. For this task, we introduce the hyper-hypernetwork framework, in which an outer hypernetwork $q$ determines the weights of an inner hypernetwork $f$, which determines the weights of the primary network $g$. 

Our experiments demonstrate the success of the trained models in producing solutions that comply with the geometric constraints and achieve the desired radiation pattern. We demonstrate that both the hypernetwork $f$ and the refinement network $t$ are required for the design of a single antenna and that the method outperforms the baseline methods. In the case of multiple antennas, the hyperhypernetwork, which consists of networks q, f, g, outperforms the baseline methods on a realistic synthetic dataset. Furthermore, it is able to predict the structure of real-world antenna designs \cite{chen_design_2018,singh_design_2016} and to suggest an alternative design that has improved array directivity for the iPhone 11 Pro Max.

\section{Related Work}
\label{sec:prev}

\citet{misilmani_machine_2019} survey design methods for large antennas, i.e., antennas the size of $\lambda/2 - \lambda/4$, where $\lambda$ is the corresponding wavelength of their center frequency. Most of the works surveyed are either genetic algorithms \cite{geneticalgo} or SVM based classifiers \cite{svm_antenna}. None of the surveyed methods incorporates geometrical constraints, which are crucial for the design of the small antennas we study, due to physical constraints.

A limited number of attempts were made in the automatic design of small antennas, usually defined by a scale that is smaller than $\lambda/10$ \cite{bulus2014center}. 
\citet{Hornby2006NASA,geneticalgo} employ genetic algorithms to obtain the target gain. \citet{Military} employ hierarchical Bayesian optimization with genetic algorithms to design an electrically small antenna and show that the design obtained outperforms classical man-made antennas.  
None of these methods employ geometric constraints, making them unsuitable for real-world applications. They also require running the antenna simulation over and over again during the optimization process. Our method requires a one-time investment in creating a training dataset, after which the design process itself is very efficient. 

A hypernetwork scheme~\cite{ha2016hypernetworks} is often used to learn dynamic networks that can adjust to the input~\cite{bertinetto2016learning,Oswald2020Continual} through multiplicative interactions~\cite{jayakumar2020multiplicative}. It contains two networks, the hypernetwork $f$, and the primary network $g$. The weights $\theta_g$ of $g$ are generated by $f$ based on $f$'s input. 
We use a hypernetwork to recover the structure of the antenna in 3D. Hypernetworks were recently used to obtain state of the art results in 3D reconstruction from a single image~\cite{Littwin_2019_ICCV}. 

We present multiple innovations when applying hypernetworks. First, we are the first, as far as we can ascertain, to apply hypernetworks to complex manufacturing design problems. 
Second, we present the concept of a hyperhypernetwork, in which a hypernetwork provides the weights of another hypernetwork. Third, we present a proper way to initialize hyperhypernetworks, as well as a heuristic for selecting which network weights should be learned as conventional parameters and which as part of the dynamic scheme offered by hypernetworks.

\section{Single Antenna Design}
\label{single_antenna}
Given the geometry of an antenna, i.e. the metal structure, one can use a Finite Difference Time Domain (FDTD) software, such as OpenEMS FDTD engine~\cite{openEMS}, to obtain the antenna's radiation pattern in spherical coordinates $(\theta,\phi)$. Applying such software to this problem, under the setting we study, has a runtime of 30 minutes per sample, making it too slow to support an efficient search for a geometry given the desired radiation pattern, i.e., solve the inverse problem.  Additionally, since it is non-differentiable, its usage for optimizing the geometry is limited.

Therefore, although our goal is to solve the inverse problem, we first build a simulation network $h$. This network is used to support a loss term that validates the obtained geometry, and to propagate gradients through this loss. The simulation network $h$ is given two inputs (i) the scale in terms of wavelength $S$ and (ii) a 3D voxel-based description of the spatial structure of the metals $V$. $h$ returns a 2D map $U$ describing the far-field radiation pattern, i.e., $U = h(S,V)$.

Specifically, $S\in \mathbb{R}^3$ specifies the arena limits. This is given as the size of the bounding box of the metal structure, in units of the wavelength $\lambda$ corresponding to the center frequency. $V$ is a voxel grid of size $64\times 64 \times 16$, which is sampled within the 3D bounding box dimensions provided by $S$. In other words, it represents a uniform sampling on a grid with cells of size $[S_1/64,S_2/64,S_3/16]$. The lower resolution along the $z$ axis stems from the reduced depth of many mobile devices. Each voxel contains a binary value: 0 for nonmetallic materials, 1 for conducting metal. The output tensor is a 2D ``image'' ${U(\theta,\phi)}$, sampled on a grid of size $64\times 64$, each covering a region of $\pi/64\times 2\pi/64$ squared arc lengths. The value in each grid point denotes the radiation power in this direction. 


\noindent The directivity gain $D=\tN(U)$ is a normalized version of $U$:
\begin{equation}
    D(\theta,\phi) = \frac{U(\theta,\phi)}{ \int_{\phi = 0}^{2\pi}  \int_{\theta = 0}^{\pi} U(\theta,\phi)\sin(\theta) \,d\theta \,d\phi }\label{eq:D}
\end{equation}

The design network solves the inverse problem, i.e., map from the required antenna's directivity gain $D$ to a representation of the 3D volume $V$. We employ a hypernetwork scheme, in which an hypernetwork $f$ receives the design parameters $D$ and $S$ and returns the weights of an occupancy network $g$. $g:[0,1]^3\rightarrow [0,1]$ is a multi-layered perceptron (MLP) that maps a point $p$ in 3D, given in a coordinate system in which each dimension of the bounding box $S$ is between zero and one, into the probability $o$ of a metallic material at point $p$. 
\begin{equation}
    \theta_g = f(D,S)\,,\quad o = g(p;\theta_g)
\end{equation} 
The weights of the hypernetwork $f$ are learned, while the weights of the primary network $g$ are obtained as the output of $f$. Therefore $g$, which encodes a 3D shape, is dynamic and changes based on the input to $f$. 

To obtain an initial antenna design in voxel space, we sample the structure defined by the functional $g$ along a grid of size $64\times 64\times 16$ and obtain a 3D tensor $O$. However, this output was obtained without considering an additional design constraint that specifies unmovable metal regions.

To address this constraint, we introduce network $t$. We denote the fixed metallic regions by the tensor $M\in \mathbb{R}^{64\times 64\times 16}$, which resides in the same voxel space as $V$.
$t$ acts on $M,O$ and returns the desired metallic structure $\bar V$, i.e., $V = t(M,O)$. 

\paragraph{Learning Objectives}
For each network, a different loss function is derived according to its nature. Since the directivity gain map is smooth with regions of peaks and nulls, the multiscale SSIM~\cite{ZWang} (with a window size of three) is used to define the loss of $h$. Let $U^*$ be the ground truth radiation pattern, which is a 2D image, with one value for every angle $\theta,\phi$. The loss of the simulation network is given by
\begin{equation}
L_h = -\text{msSSIM}(U^*,h(S,V))
\end{equation}
The simulation network $h$ is trained first before the other networks are trained.

The loss of the hypernetwork $f$ is defined only through the output of network $g$ and it backpropagates to $f$.
\begin{equation}
\label{eq:LG}
L_g = \text{CrossEntropy}(g(p,f(D,S)),y_p)
\end{equation}
where $y_p$ is the target metal structure at point $p$. This loss is accumulated for all samples on a dense grid of points $p$.

For $t$, the multitask paradigm~\cite{kendall2018multitask} is used, in which the balancing weights $\alpha_i$ are optimized as part of the learning process.
\begin{equation}
\label{eq:multiloss}
multiloss([l_1 ...l_n]^{T}) = \sum_{i\in [1,n]} exp(-\alpha_i)\cdot l_i + \alpha_i
\end{equation} 
\noindent where $l_i$ are individual loss terms and $\alpha\in\mathbb{R}^n$ is a vector of learned parameters. Specifically, $t$ is trained with the loss $L_t = multiloss(L_{OBCE},L_{msSSIM})$ for
\begin{align}
L_{OBCE} &= -\frac{1}{|M_p|}\sum_{p\in \{M_p\}} M_p\cdot log(\bar V_p) \label{eq:LBCE}\\
    L_{msSSIM} &= -msSSIM(\tN(h(S,\bar V)),D)\,.
    \end{align}

The first loss $L_{OBCE}$ is the binary cross entropy loss that considers only the regions that are marked by the metallic constraint mask $M$ as regions that must contain metal. The second loss $L_{msSSIM}$ is the SSIM of the radiation patterns ($\tN$ is the normalization operator of Eq.~\ref{eq:D}).

\begin{figure}
  \begin{tabular}{@{}c@{}}
     \includegraphics[width=1\linewidth]{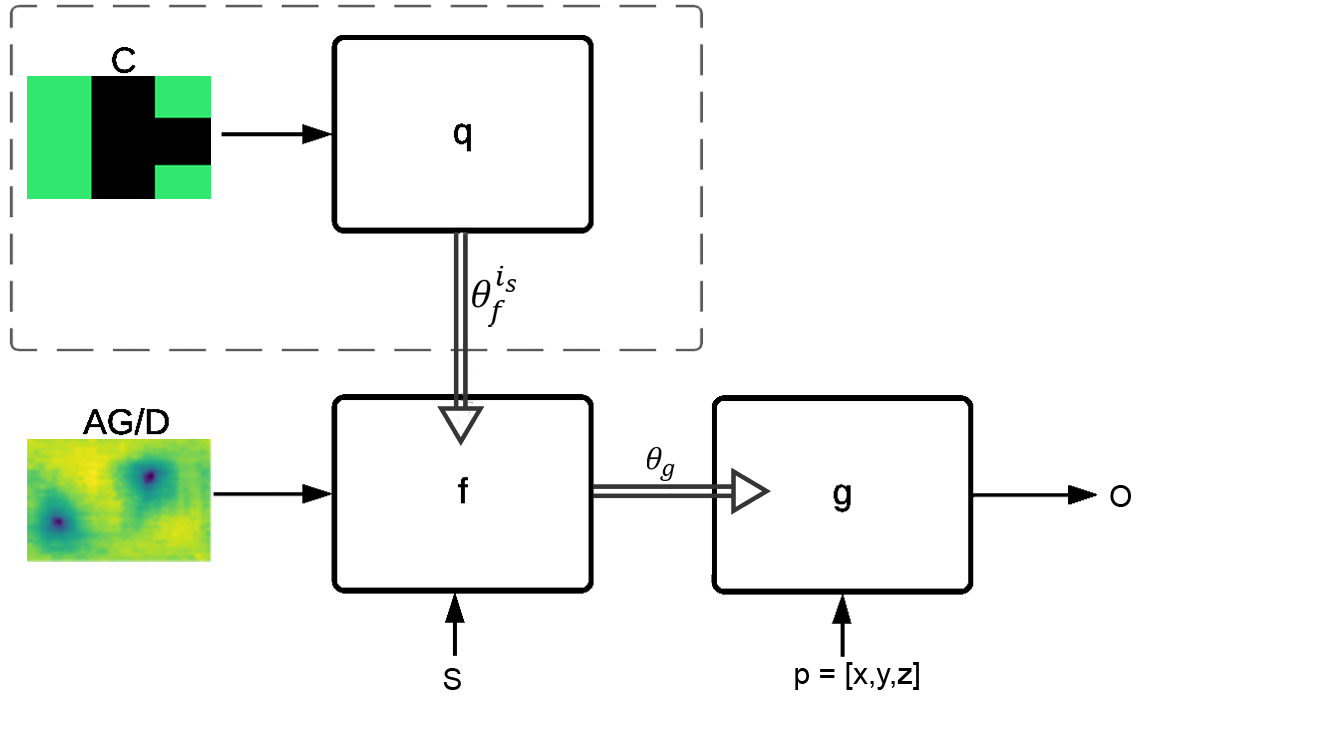} \\
  \end{tabular}
    \caption{The architecture of the hyperhypernetwork $q$, the hypernetwork $f$, and the primary network $g$ used in the antenna array design. In the case of the single antenna design, only $f,q$ are used. The hyperhypernetwork $q$ is given the constraint plane $C$ and outputs the parameters $\theta_f$ of network $f$.  The hypernetwork $f$, given the bounding box $S$ and target directivity $D$ or array gain $AG$, produces the weights $\theta_g$ of network $g$. The primary network $g$ maps a point $p$ in 3D into the probability of metal occupancy at that point. The simulation network $h$ used to compute the loss and the refinement network $t$ are not shown in the diagram.}
    \label{fig:farch}
\end{figure}
\begin{figure}[t]
    \centering
    \begin{tabular}{c}
    \includegraphics[trim={0 0 0 0},width=0.45\textwidth]{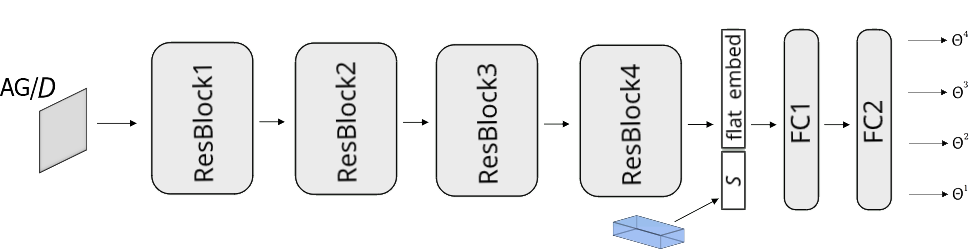}\\
    (a)\\
    \includegraphics[trim={0 0 0 0},width=0.45\textwidth]{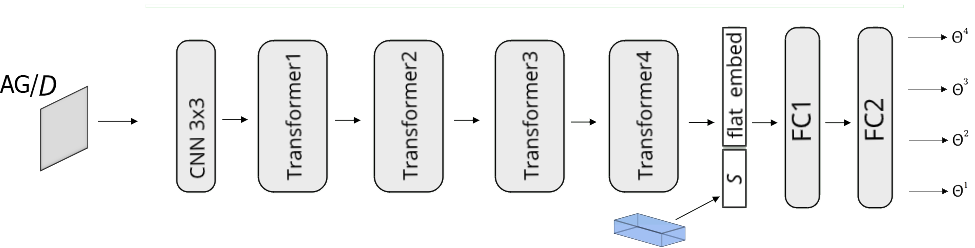}\\
    (b)\\
    \end{tabular}
    \caption{The two variants of network f. (a) The ResNet variant. (b) The Transformer variant.}
    \label{fig:farchbothoptions}
\end{figure}
\section{Antenna Array}

Antenna arrays are used in the current design of consumer communication devices (and elsewhere) to enable complex radiation patterns that are challenging to achieve by a single antenna. For example, a single cellular device needs to transmit and receive radio frequencies with multiple wifi and cellular antennas ~\cite{Bogale_massive_mimo}. 

We present a method for designing antenna arrays that is based on a new type of hypernetwork. For practical reasons, we focus on up to $N_a=6$ antennas. See appendix for the reasoning behind this maximal number.

While a single antenna is designed based on a target directivity $D$, an array is defined based on a target array gain $AG$. Assuming, without loss of generality, that we consider a beamforming direction of zero, $AG$ is defined, for the observation angle $\phi$, $\theta$ as
\begin{equation}
     AG(\theta,\phi) =  \sum_{ant}U_{ant}(\theta,\phi)\exp(-jkr_{ant})
     \label{eq:AG}
\end{equation}
where $U_{ant}(\theta,\phi) \in\mathbb{R}$ is the real valued radiation pattern of single element in the array, $k$ is the wave vector, defined as
\begin{equation}
    k = 2\pi/\lambda \times[sin(\theta)cos(\phi),sin(\theta)sin(\phi),cos(\theta)]
\end{equation}
and $r_{ant} = [r_{x},r_{y},r_{z}] \in \mathbb{R}^3$ is the element center phase position in 3D coordinates.


In addition to the array gain pattern $AG$ and the global 3D bounding box $S$, antenna arrays adhere to physical design constraints. For example, for mobile devices, multiple areas exists where antenna elements can not be fitted, due to electromagnetic interference, product design, and regulations~\cite{fcc_sar}.  

Unlike the single antenna case, which is embedded in a 3D electrical board, which is captured by the constraint tensor $M$, multiple antennas are mostly placed outside such boards. We, therefore, assume that the constraints are independent of the $z$ and formulate these as a binary matrix $C$.
\begin{equation}
  C(x,y)=\left\{
  \begin{array}{@{}ll@{}}
    1, & \text{if}\ (x,y)\not\in AP \\
    0, & \text{otherwise}
  \end{array}\right.
  \label{eq:constraint}
\end{equation}
Where $AP$ is the subset of positions in the XY plane in which one is allowed to position metal objects.



{\bf Hyperhypernetworks\quad} We introduce a hyperhypernetwork scheme, which given high-level constraints on the design output, determines the parameters of an inner hypernetwork that, in return, outputs those of a primary network.

{Let $q$ be the hyperhypernetwork, and let $i_s$ be the indices of the subset of the parameters $\theta_f$ of network $f$. $q$ returns this subset of weights $\theta^{i_s}_{f}$,  based on the constraint matrix $C$:
\begin{equation} 
\theta^{i_s}_{f}  = q(C;\theta_{q})\
\end{equation}
where $\theta_{q}$ are learned parameters. Below, ${i_c}$ indexes parameters of $f$ and is complementary to $i_s$. The associated weights $\theta^{i_c}_{f}$ are learned conventionally and are independent of $C$.

Network $f$ gets as input ${AG,S}$ (defined in Eq.~\ref{eq:AG} and Sec.~\ref{single_antenna}, respectively). The former is given in the form of a 2D tensor sampled on a grid of size $64\times64$, each covering a region of $\pi/64\times2\pi/64$ squared arc lengths, and the latter is in $\mathbb{R}^3$.  The output of $f$ are the weights of network $g$, $\theta_g$.
\begin{equation}
    \theta_{g} = f(AG,S;\theta^{i_s}_{f},\theta^{i_c}_{f})
\end{equation}

The output of the primary network $g$ given a 3D coordinate vector $p \in \mathbb{R}^3$ is a tuple $O = g(p;\theta_{g})$, 
where $O=[Pr(p\in Metal),Pr(p\in Valid Antenna)] $ is concatenation of two probabilities, the probability of 3D point $p$ being classified as metal voxel or a dielectric voxel, and the probability of this point belongs to a valid antenna structure.} 
The high-level architecture, including the networks $q,f,g$ is depicted in Fig.~\ref{fig:farch} and the specific architectures are given in Sec.~\ref{sec:arch}.

Unlike the single antenna case, where the metallic constraints are tighter, for the antenna array, we do not employ a refinement network $t$. The training loss is integrated along points $p$ in 3D and is a multiloss (Eq.~\ref{eq:multiloss}) of a structural and a constraint loss, similar to the single antenna. The structural loss is given by $L_{s} = \sum_p CrossEntropy(O_1[p], y_{p})$, where $y_p$ is the target metal structure at point p and $O_k$ is the $k$th element of $O$. The constraint loss is $L_{C} = -\frac{1}{|\{C_{p}\}|}\sum_{p\in \{C_p\}} C_p\cdot log(1-O_{2{[p]}})$, where $(p_x,p_y)$ are the X and Y coordinates of the input point $p$.





{\bf Initialization\quad} Hypernetworks 
present challenges with regards to the number of learned parameters and are also challenging to initialize~\cite{Littwin_2019_ICCV,Chang2020Principled}. Below, we (i) generalize the initialization scheme of \cite{Chang2020Principled} to the hyperhypernetwork, and (ii) propose a way to select which subset of parameters $i_{s}$ would be determined by the network $q$, and which would be learned conventionally as fixed parameters that do not depend on the input. 

Define the hypernwetwork $f=f_{n}\circ \dots \circ f_2 \circ f_1$ as a composition of $n$ layers. Similarly we define the hyperhypernetwork $q=q_{n}\circ \dots \circ q_2 \circ q_1$ as a composition of $n$ layers, and define $w^{(c)}_{j} = q_{j}\circ \dots \circ q_{1}(c)$, where $c$ is the hyperhypernetwork input.

We assume that $i_s$, the set of parameters that are determined by $q$ contains complete layers, which we denote as $j_s \subset [n]$.
The computation of $f_j$ on the embedding $e$,
\begin{equation}
    f_j(e)=\left\{
  \begin{array}{@{}ll@{}}
    f_j(e), & \text{if}\ j\not\in j_s \\
    (q_{n}^{j}\times w^{(c)}_{n-1}) e, & j\in j_s
  \end{array}\right.
\end{equation}
Where $\times$ denotes tensor multiplication along the relevant dimension and $q_{n}^{j}$ is the portion of the hyperhypernetwork last layer corresponds the j-th layer of $f$.

For $f_j$ where $j\not\in j_s$ we use \cite{Chang2020Principled} results for initialization.
We initialize $q_{n-1},\dots,q_1$ using the Xavier fan in assumption, obtaining $Var(w^{(c)}_{j}) = Var(c)$. 
The variance of the output of the primary network, denoted $y$, given primary network input $x$, hypernetwork input $e$, and hyperhypernetwork input $c$ is
\begin{equation}
\label{eq:varexp}
\begin{split}
    Var(y) &= \sum_j\sum_k Var(f_{n}[j,k])Var(f_k(e))Var(x_j) \\ &=\sum_j(\sum_{k\not\in j_s} Var(f_{n}[j,k])Var(f_k(e))Var(x_j)\\
    &+\sum_{k\in j_s}\sum_m (Var(q_{n}[k,m])Var(w^{(c)}_{j}[m])Var(e)\\
    &Var(f_{n}[j,k])Var(x_j)))
\end{split}
\end{equation}
where we use brackets to index matrix or vector elements. We propose to initialize elements $k$ of the last layer of $q$ as
\begin{equation}
    Var(q_{n}[k]) = ({d_m Var(c)})^{-1}
\end{equation}
where $d_m$  is the fan-in of the last hyperhypernetwork layer $q_{n}$. This way we obtain the desired
\begin{multline}
\label{eq:finalinitialzation}
Var(y) = Var(x_j)\frac{d_k-Q}{d_k} + \sum_{j,k\in j_s,m} \frac{1}{d_k d_m d_j} Var(x_j)\\ = Var(x_j)\frac{d_k-Q}{d_k} +\frac{Q}{d_k}Var(x_j) = Var(x_j)
\end{multline}
where $Q = |i_s|$ is the number of parameters that vary dynamically as the output of the hyperhypernetwork.



{\bf Block selection\quad} Since the size of network $q$ scales with the number $Q$ of parameters in $f$ it determines, we limit, in most experiments, $Q<10,000$. This way, we maintain the batch size we use despite the limits of memory size. 
The set of parameters $i_s$ is selected heuristically as detailed below.

Let $n$ be the number of layers in network $f$. We arrange the parameter vector $\theta_f$ by layer, where $\theta_f^j$ denotes the weights of a single layer $\theta_f = [\theta_f^0,...,\theta_f^{n-1}]$ 
and $|\theta_f^j|$ is the number of parameters in layer $j$ of $f$. The layers are ordered by the relative contribution of each parameter, which is estimated heuristically per-layer as a score $H_j$

$H_j$ is computed based on the distribution of losses on the training set that is obtained when fixing all layers, except for layer $j$, to random initialization values, and re-sampling the random weights of layer $j$ multiple times. This process is repeated $10,000$ times and the obtained loss values are aggregated into $1,000$ equally spaced bins. The entropy of the resulting $1,000$ values of this histogram is taken as the value $H_j$. Since random weights are used, this process is efficient, despite the high number of repetitions.

The method selects, using the Knapsack algorithm \cite{dantzig_1955},  a subset of the layers with the highest sum of $H_j$ values such that the total number of parameters (the sum of $|\theta_f^j|$ over $j_s$) is less than the total quota of parameters $Q$.

\section{Architecture}
\label{sec:arch}

The network $h$, which is used to generate additional training data and to backpropagate the loss, consists of a CNN ($3\times3$ kernels) applied to $V$, followed by three ResNet Blocks, a concatenation of $S$, and a fully connected layer. ELU activations~\cite{clevert2015fast} are used.

The primary network 
$g$ is a four layer MLP, each with 64 hidden neurons and ELU activations,  except for the last activation, which is a sigmoid (to produce a value between zero and one). The MLP parametrization, similarly to \cite{Littwin_2019_ICCV}, is given by separating the weights and the scales, where each layer $j$ with input $x$ performs $(x^\top\theta_w^j)\cdot \theta^j_s + \theta^j_b$, 
where $\theta^j_w\in \mathbb{R}^{d_1\times d_2},\theta^j_s\in \mathbb{R}^{d_2}$,and $\theta^j_b\in\mathbb{R}^{d_2}$ are the weight-, scale- and bias-parameters of each layer, respectively, and $\cdot$ is the element-wise multiplication. The dimensions are $d_1=3$ for the first layer, $d_1=64$ for the rest, and $d_2=64$ for all layers, except the last one, where it is one.

For the hypernetwork $f$, we experiment with two designs, as shown in Fig.~\ref{fig:farchbothoptions}. Architecture (a) is based on a ResNet and architecture (b) has a Transformer encoder \cite{vaswani2017attention} at its core. Design (a) $f$ has four ResNet blocks, and two fully connected (FC) layers. $D$ propagates through the ResNet blocks and flattened into a vector. $S$ is concatenated to this vector, and the result propagates through two FC layers. The weights of the last Linear unit in $f$ are initialized, according to \cite{Chang2020Principled}, to mitigate vanishing gradients.

Design (b) of the hypernetwork $f$ consists of three parts: (1) a CNN layer that is applied to $AG$, with a kernel size of $3\times3$. (2) a Transformer-encoder that is applied to the vector of activations that the CNN layer outputs, consisting of four layers each containing: (i) multiheaded self-attention, (ii) a fully connected layer. The self attention head was supplemented with fixed sine positional encoding \cite{parmar_image_2018}. (3) two fully connected (FC) layers, with an ELU activation in-between. The bounding box $S$ is concatenated to the embedding provided by the transformer-encoder, before it is passed to the FC layers.

The hyperhypernetwork $q$ is a CNN consisting of four layers, with $ELU$ as activations and a fully connected layer on top. The input for $q$ is the constraint image $C$, of size $192\times128$, divided to a grid of $2\times3$ cells ($64\times 64$ regions), denoting the possible position of the individual antennas. The network $q$ outputs are the selected weights $\theta_f^{i_s}$ of $f$.

The network $t$, which is used to address the metallic constraints in the single antenna case, consists of three ResNets $m_1,m_2,m_3$ and a vector $w\in\mathbb{R}^2$ of learned parameters. It mixes, using the weights $w$, the initial label $O$ and the one obtained from the sub-networks: $T_1 = m_1(M), T_2 = m_2(O), T_3 = m_3([T_1,T_2]), \bar V = w_1 O + w_2 T_3$, 
where $T_1,T_2,T_3\in \mathbb{R}^{64 \times 64 \times 16}$, and $[T_1,T_2]$ stands for the concatenation of two tensors, along the third dimension. Both $m_1,m_2$ are ResNets with three blocks of 16 channels. $m_3$ is a single block ResNet with 32 channels.


%

\begin{table*}[t]
  \caption{The performance obtained by our method, as well as the baseline methods and ablation variants. See text for details.} 
    \label{tab:results}
\centering
 \begin{tabular}{lcccc} 
 \toprule
 & \multicolumn{2}{c}{Radiation pattern} & \multicolumn{2}{c}{3D shape}\\
 \cmidrule(lr){2-3}
 \cmidrule(lr){4-5}
Method & MS-SSIM $\uparrow$  & SNR[dB] $\uparrow$ &  IOU $\uparrow$ & $M$-Recall $\uparrow$\\
 \midrule
(i) Baseline nearest neighbor &0.88 $\pm$ 0.06& 32.00 $\pm$ 0.40 &0.80 $\pm$ 0.11 & 0.05 $\pm$ 0.03 \\ 
(ii) Baseline nearest neighbor under metallic constraints &0.89 $\pm$ 0.09& 32.30 $\pm$ 0.30 &0.79 $\pm$ 0.13 & 0.89 $\pm$ 0.07 \\ 
(ours ResNet variant)  $\bar V=t(M,O)$& {\bf 0.96} $\pm$ 0.03 & 36.60 $\pm$ 0.45 & 0.86 $\pm$ 0.09& {\bf 0.96} $\pm$ 0.01\\
(ours Transformer variant)  $\bar V=t(M,O)$& {\bf 0.96} $\pm$ 0.04 & {\bf 36.62} $\pm$ 0.52 & {\bf 0.88} $\pm$ 0.12& 0.95 $\pm$ 0.02\\ 
 \midrule
(iii.a) No refinement ResNet variant $V=g(p,\theta_{D,S})$ & 0.91 $\pm$ 0.05 & 32.80 $\pm$ 0.41 & 0.84 $\pm$ 0.08& 0.06 $\pm$ 0.03 \\
(iii.b) No refinement Transformer variant $V=g(p,\theta_{D,S})$ & 0.93 $\pm$ 0.02 & 34.80 $\pm$ 0.60 & 0.86 $\pm$ 0.11& 0.04 $\pm$ 0.02 \\
(iv.a) No hypernetwork ResNet variant (training $t$ with $f'$) &0.78 $\pm$ 0.12& 22.90 $\pm$ 1.60 &0.81 $\pm$ 0.11 & 0.90 $\pm$ 0.05 \\
(iv.b) No hypernetwork Transformer variant (training $t$ with $f'$) &0.75 $\pm$ 0.17& 21.30 $\pm$ 2.20 &0.79 $\pm$ 0.11 & 0.90 $\pm$ 0.05 \\
(v.a) ResNet variant,$t$ is trained using a structure loss &0.92 $\pm$ 0.07& 33.00 $\pm$ 0.55 &0.84 $\pm$ 0.09 & 0.91 $\pm$ 0.06 \\ 
(v.b) Transformer variant, $t$ is trained using a structure loss &{0.94} $\pm$ 0.04& {33.70} $\pm$ 0.55 &{0.87 }$\pm$ 0.04 & {\bf0.96} $\pm$ 0.04\\ 
 \bottomrule
\end{tabular}
\end{table*}

\begin{figure}[t]
  \begin{tabular}{@{}c@{~}c@{}}
     \includegraphics[width=0.483\linewidth]{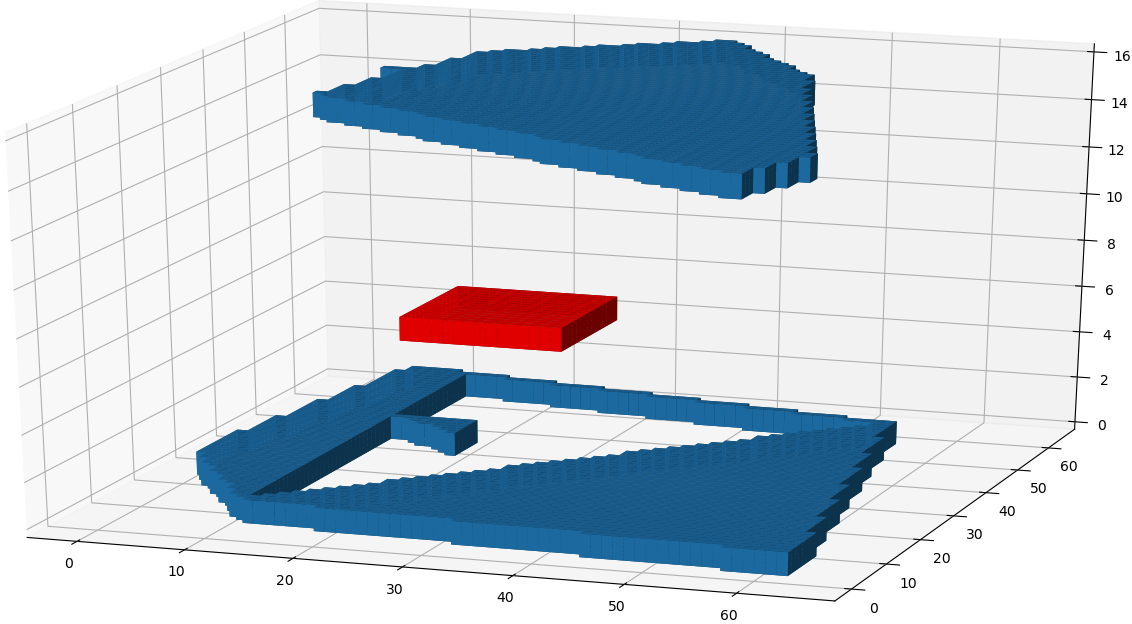} & 
     \includegraphics[width=0.483\linewidth]{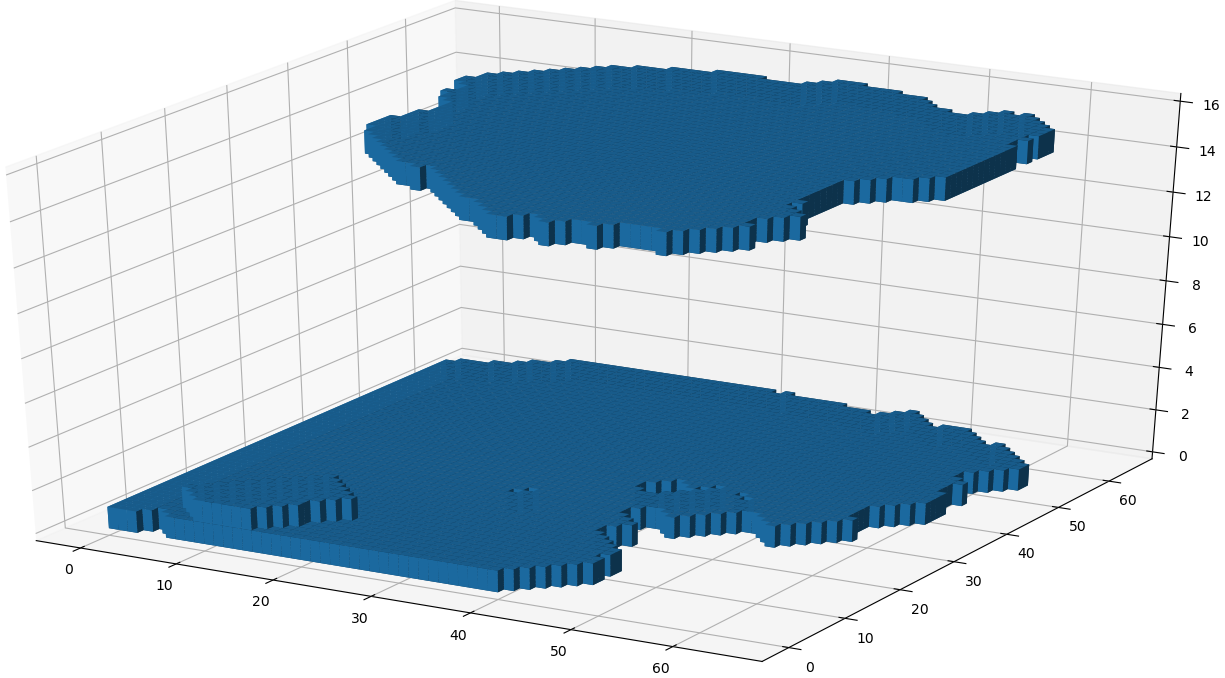} \\
     (a)&(b)\\
     \includegraphics[width=0.483\linewidth]{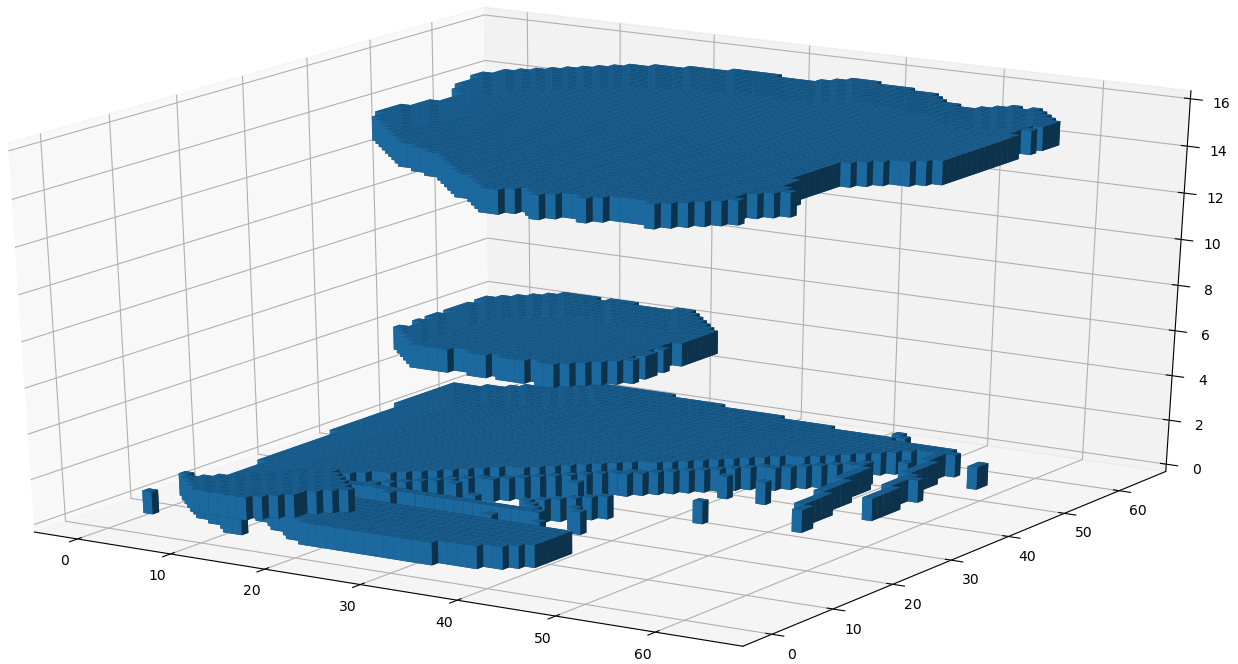} &
     \includegraphics[width=0.483\linewidth]{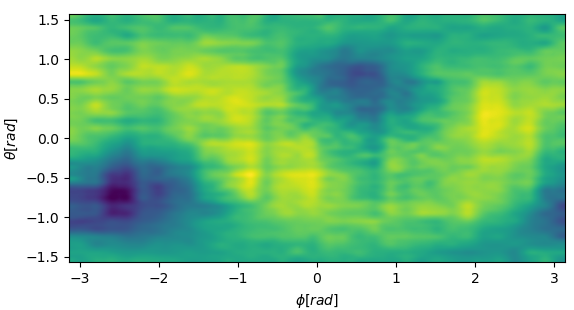}\\
     (c)&(d)\\
     \includegraphics[width=0.483\linewidth]{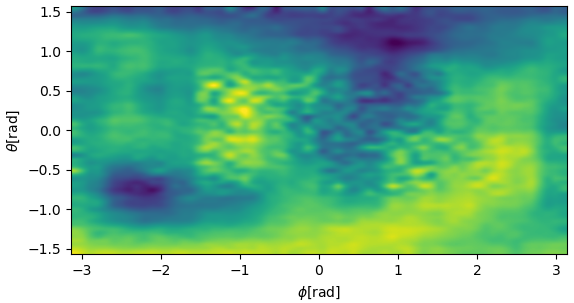} &

     \includegraphics[width=0.483\linewidth]{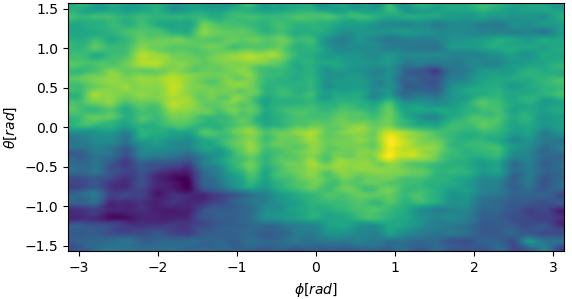} \\
     (e)&(f)\\
  \end{tabular}
    \caption{A typical test-set instance. (a) The ground truth structure $V$, with the metallic constraint regions $M$ marked in red. (b) The output structure $g(\cdot,\theta_{D,S})$ of the hypernetwork. (c) The output $\bar V$ of the complete method. (d-f) The directivity gain of $V$, $g(\cdot,\theta_{D,S})$, and $\bar V$, respectively.}
    \label{fig:sample}
\end{figure}

\begin{figure}[t]
  \begin{tabular}{@{}c@{~}c@{}}
     \includegraphics[width=0.4832\linewidth,height=0.46\linewidth]{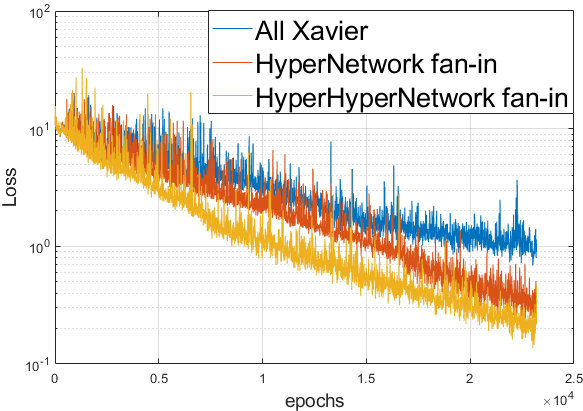} &
     \includegraphics[width=0.4832\linewidth,height=0.46\linewidth]{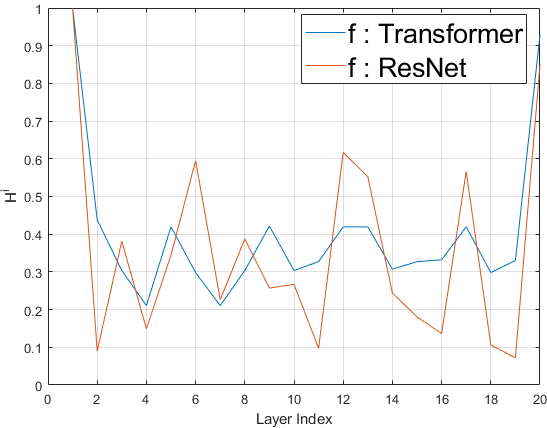}\\
     (a) & (b)\\
  \end{tabular}
    \caption{(a) Loss per epochs for the different initialization scheme of $q$ (ResNet $f$) ,Transformer in appendix. (b) The mean per-layer score obtained for the entropy based selection heuristic. the selected layers are [1,9,16] (ResNet) and [1,2,15] (Transformer).}
    \label{fig:initial}
\end{figure}
\begin{figure}[t]
  \begin{tabular}{@{}c@{~}c@{}}
     \includegraphics[width=0.4830255\linewidth]{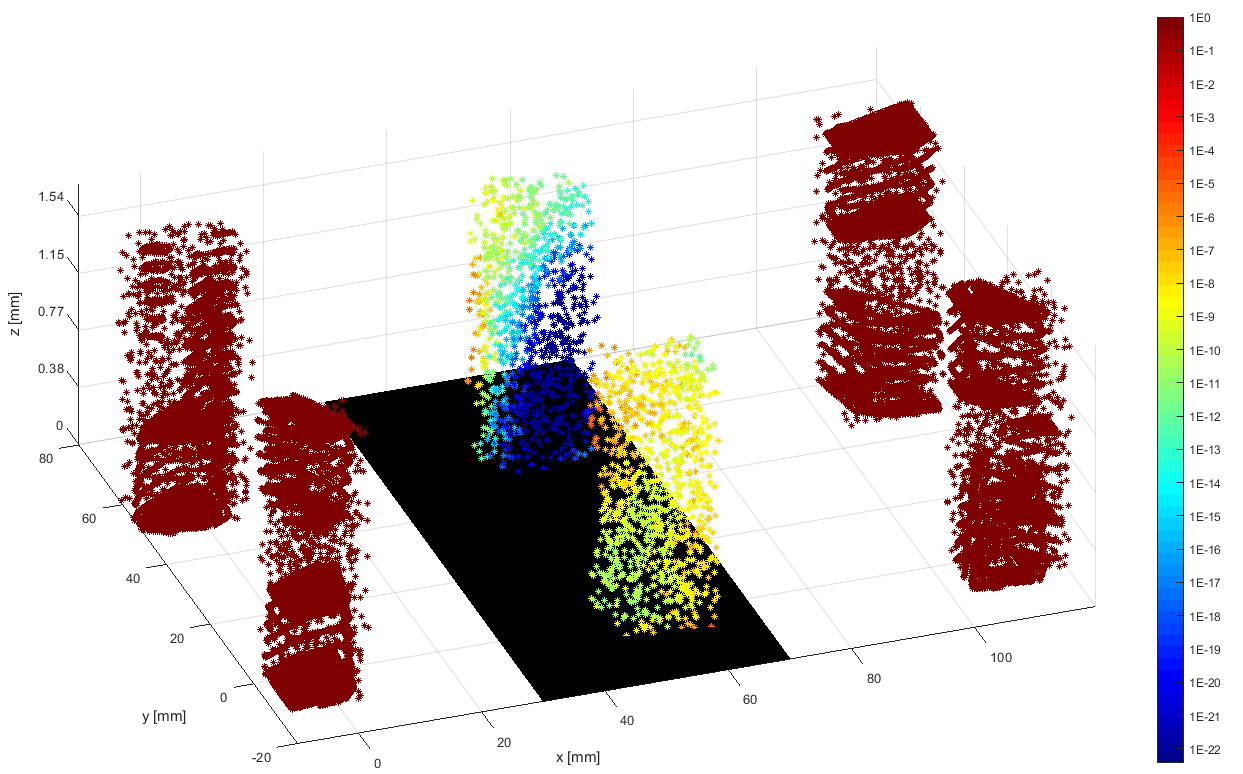} & 
     \includegraphics[width=0.4830255\linewidth]{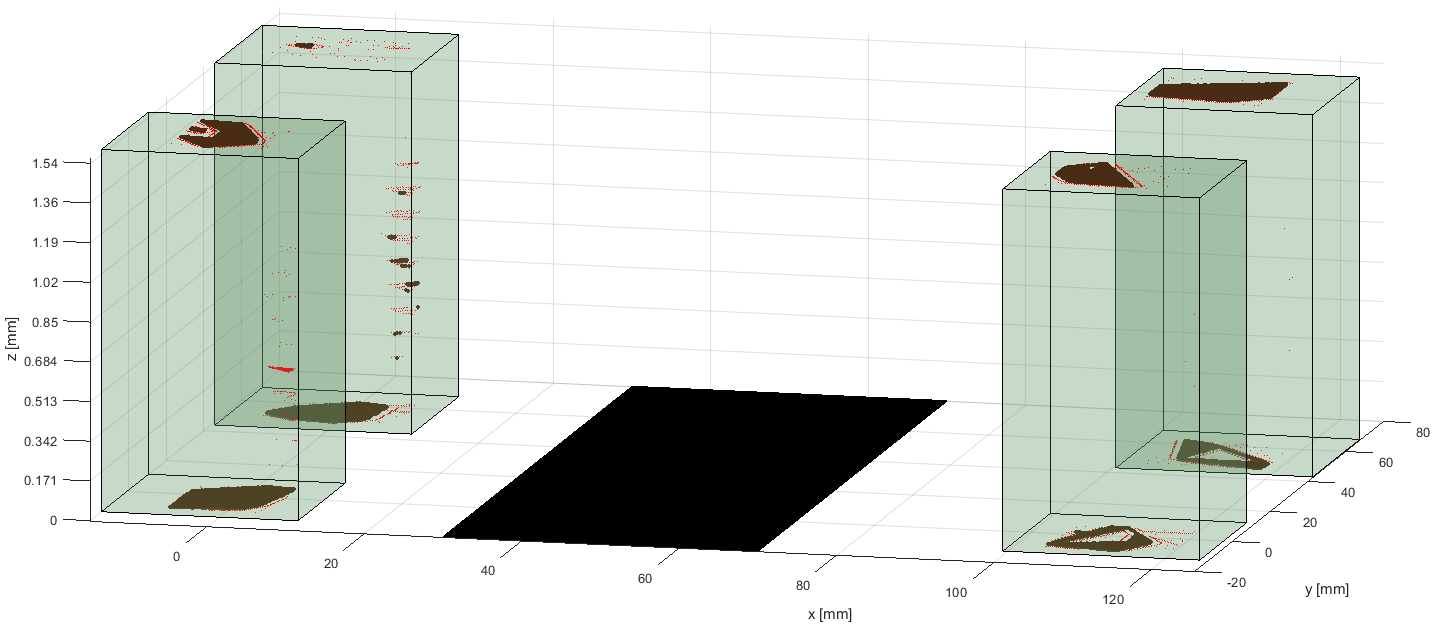} \\ 
          (a) & (b)\\
     \includegraphics[width=0.4830255\linewidth]{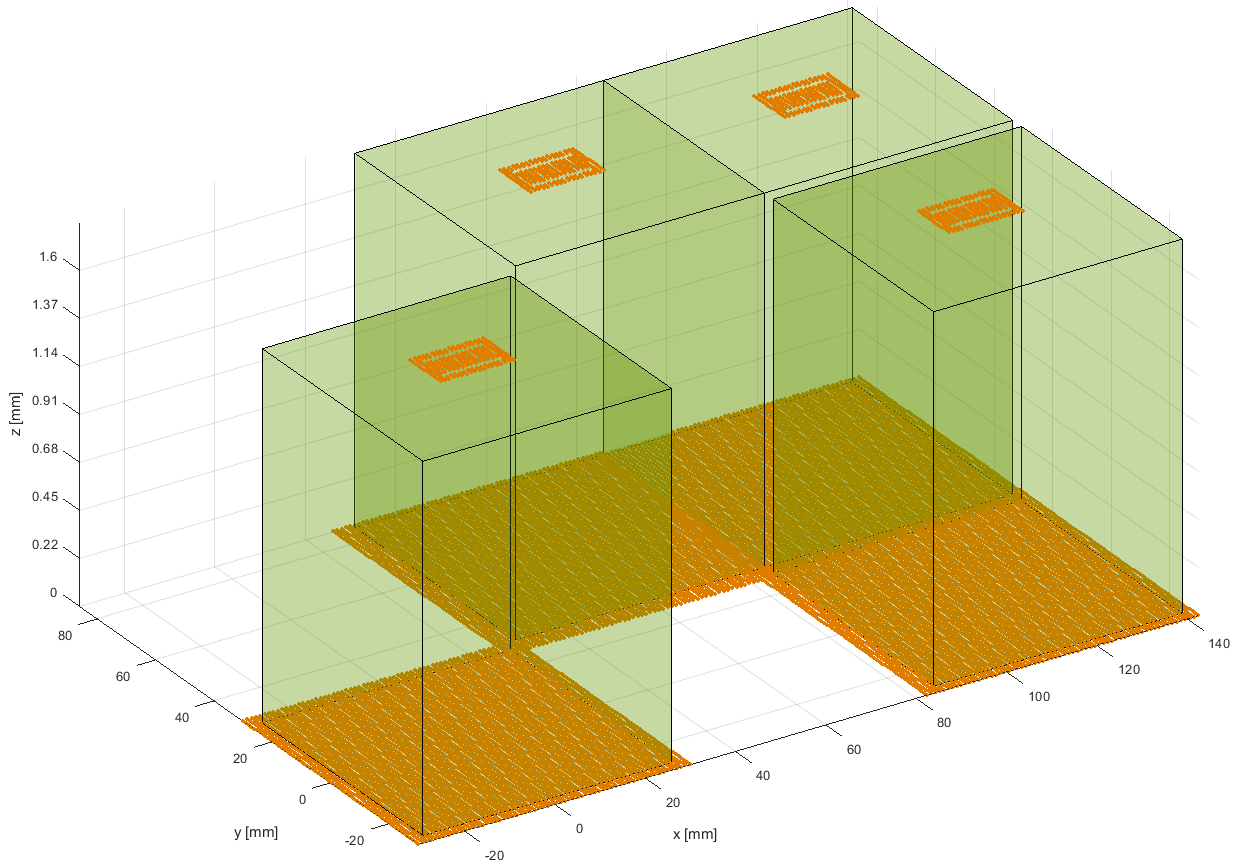} &
     \includegraphics[width=0.4830255\linewidth]{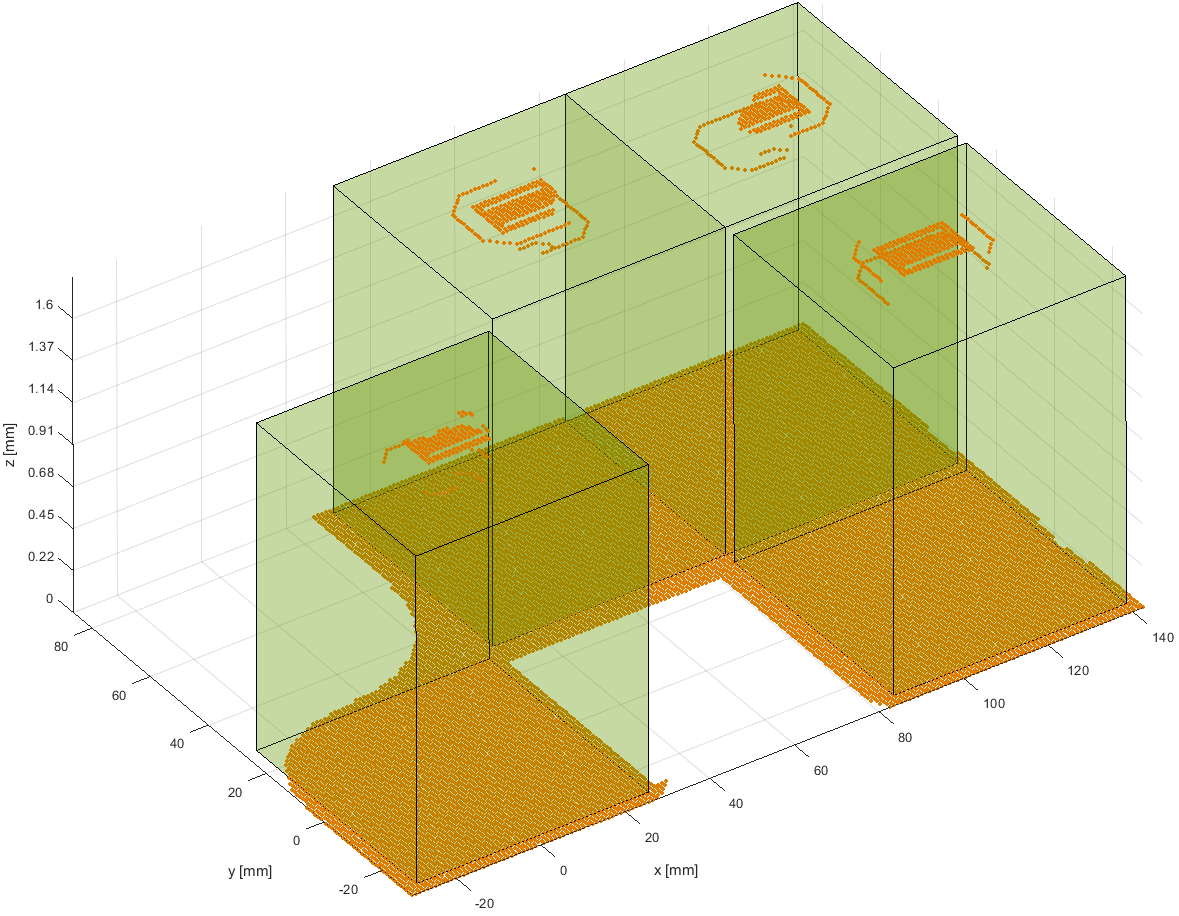} \\
     (c) & (d)\\
  \end{tabular}
    \caption{(a) The probability of points belong to a valid antenna in a synthetic test instance. The constraint plane is marked as black. (b) Same sample, the regions correctly classified as antenna are marked in brown, misclassified is marked in red.  (c) The ground truth of a slotted antenna array. (d) Our network design. See appendix for ablations.}
    \label{fig:array_sample}
\end{figure}

\begin{table}[t]
  \caption{The performance for designing antenna arrays. $C$-ratio is the fraction of the antenna volume that complies with $C$.}
    \label{tab:results_array}
\centering
 \begin{tabular}{@{}l@{~}c@{~~}c@{}} 
 \toprule
Method & IOU $\uparrow$ & $C$-ratio $\uparrow$\\
\midrule
Nearest neighbor baseline &0.48 $\pm$ 0.07& 0.25 $\pm$ 0.08 \\ 
(ours ResNet version, $Q=10^4$) &0.86 $\pm$ 0.03& 0.79$\pm$ 0.03\\
(ours ResNet version, $Q=\inf$) &0.88 $\pm$ 0.05& 0.80$\pm$ 0.04\\
{(ours Transformer ver, $Q=10^4$)}   & {\bf 0.93} $\pm$ 0.01 & 
{\bf 1.0} $\pm$ 0.0\\
(ours Transformer ver, $Q=\inf$) &{\bf 0.93} $\pm$ 0.02& {\bf 1.0}$\pm$ 0.01\\
\midrule
(i.a) Hypernet, ResNet $f$ &0.75 $\pm$ 0.06& 0.14$\pm$ 0.01 \\
(i.b) Hypernet, Transformer $f$ &0.85 $\pm$ 0.01& 0.18 $\pm$ 0.04 \\
(ii.a) ResNet w/o hypernet & 0.70 $\pm$ 0.06& 0.09 $\pm$ 0.03\\ 
(ii.b) Transformer w/o hypernet &0.79 $\pm$ 0.03& 0.10 $\pm$ 0.02 \\
 \bottomrule
\end{tabular}
  \caption{The performance obtained for two real-world antennas. $C$R is the fraction of the antenna volume that complies with $C$.}
    \label{tab:results_fab}
\begin{center}
 \begin{tabular}{@{}l@{~}c@{~}c@{~}c@{~~}c@{}} 
 \toprule
 & \multicolumn{2}{c}{Slotted Patch} & \multicolumn{2}{c}{Patch}\\
 \cmidrule(lr){2-3}
 \cmidrule(lr){4-5}
Method &IOU  & $C$R  &  IOU  & $C$R \\
 \midrule
Nearest neighbor  &0.57 &0.70 &0.81& 0.80  \\ 
(ours ResNet version)  &0.85 &0.91 &0.89& 0.96  \\ 
(ours ResNet version $Q=\inf$)  &0.87 &0.93 &{\bf 0.90}& 0.96  \\ 
{(ours Transformer ver)}  & {\bf 0.89} & 0.93 &{\bf0.90}&{\bf1.0}\\
(ours Transformer ver $Q=\inf$)  &0.88 &{\bf0.94} &{\bf 0.90}& {\bf 1.0}  \\ 
(i.a) Hypernet ResNet  &0.82 &0.83 &0.87& 0.94  \\
(i.b) Hypernet Transformer   &0.76 & 0.79 &0.88 & {\bf 1.0} \\ 
(ii.a) Transformer w/o hypernet  &0.63 &0.79&0.82& 0.83  \\
(ii.b) ResNet without hypernet  &0.61 & 0.75 &0.82& 0.80  \\ 
 \bottomrule
\end{tabular}
\end{center}
\end{table}

\begin{table}[t]
  \caption{The performance for the iPhone 11 Pro Max design.}
    \label{tab:results_iphone}
\centering
 \begin{tabular}{@{}l@{~}cc@{}} 
 \toprule
Method &  Directivity[dBi]  & $C$-ratio \\
 \midrule
(Apple's original design) &3.1 &1.0\\
\midrule
Nearest neighbor & 1.5 & 0.05\\
(ours ResNet ) & 4.7 &0.96\\
(ours ResNet  $Q=\inf$) & 4.7 &0.97\\
{(ours Transformer )} & {\bf 5.2} & {\bf 1.0}\\
(ours Transformer  $Q=\inf$) &  5.1 & {\bf 1.0}\\
\midrule
(i.a) Hypernet ResNet  & 2.1 & 0.68\\
(i.b) Hypernet Transformer &5.0 & 0.70\\ 
(ii.a) ResNet w/o hypernet &1.1& 0.37\\ 
(ii.b) Transformer w/o hypernet &1.8& 0.33\\
 \bottomrule
 \vspace{-7mm}
\end{tabular}
\end{table}

\section{Experiments}

We present results on both synthetic datasets and on real antenna designs. In addition we describe a sample of manufactured antenna array and its' performance in the appendix. Training, in all cases, is done on synthetic data. For all networks, the Adam optimizer~\cite{kingma2014adam} is used with a learning rate of $10^{-4}$ and a decay factor of 0.98 every 2 epochs for $2,000$ epochs, the batch size is 10 samples per mini-batch .

{\bf Single Antenna\quad}
\label{sec:expsingle}
The single antenna synthetic data is obtained at the WiFi center frequency of 2.45GHz. The dielectric slab size, permeability, and feeding geometry are fixed during all the experiments. The dataset used in our experiments consists of $3,000$ randomly generated PCB antenna structures, with a random metal polygons structure. The OpenEMS FDTD engine~\cite{openEMS} was used to obtain the far-field radiation pattern $U$. The dataset is then divided into train/test sets with a 90\%-10\% division. 

  We train the simulation network $h$ first, and then the design networks $f,t$.  For the simulating network, an average of 0.95 Multiscale-SSIM score over the validation set was achieved. Once $h$ is trained on the initial dataset, another $10^4$ samples were generated and the radiation pattern is inferred by $h$ (more efficient than simulating). When training the design networks, the weight parameters of the multiloss $L_t$ are also learned. The values obtained are $\alpha_{msSSIM} \sim 10 \alpha_{OBCE}$.

For the design problem, which is our main interest, we use multiple evaluation metrics that span both the 3D space and the domain of radiation patterns. To measure the success in obtaining the correct geometries, we employ two metrics: the IOU between the obtained geometry $\bar V$ and the ground truth one $V$, and the recall of the structure $\bar V$ when considering the ground truth metallic structure constraints $M$. The latter is simply the ratio of the volume of the intersection of $M$ and the estimated $\bar V$ over the volume of $M$.

To evaluate the compliance of the resulting design to the radiation specifications, we use either the MS-SSIM metric between the desired radiation pattern $D$ and the one obtained for the inferred structure of each method, or the SNR between the two.

We used the following baseline methods: 
 (i) nearest neighbor in the radiation space, i.e., the training sample  that maximizes the multiscale SSIM metric relative to the test target $D$, and (ii) a nearest neighbor search using the SSIM metric out of the samples that have an M-recall score of at least 85\%. In addition, we used the following ablations and variants of our methods in order to emphasize the contribution of its components:  (iii) the output of the hypernetwork, before it is refined by network $t$, (iv) an alternative architecture, in which the hypernetwork $f$ is replaced with a ResNet/Transformer-based $f'$ of the same capacity as $f$, which maps $D,S$ to O directly $O = f'(D,S)$, and  (v) the full architecture, where the loss $L_{msSSIM}$ is replaced with the cross entropy loss on $\bar V$ with respect to the ground target structure (similar to $L_g$, Eq.~\ref{eq:LG}). 
 This last variant is to verify the importance of applying a loss in the radiation domain. 
 
The results are reported in Tab.~\ref{tab:results}. As can be seen, the full method outperforms the baseline and ablation methods in terms of multiscale SSIM, which is optimized by both our method and the baseline methods. Our method also leads with respect to the SNR of the radiation patterns, and with respect to IOU. A clear advantage of the methods that incorporate the metallic surface constraints over the method that do not is apparent for the $M$-Recall score, where our method is also ranked first.

The hypernetwork ablation (iii), which does not employ $t$, performs well relative to the baselines (i,ii), and is outperformed by the ablation variant (v) that incorporates a refinement network $t$ that is trained with a similar loss to that of $f$. The difference is small with respect to the radiation metrics and IOU and is more significant, as expected, for $M$-Recall, since the refinement network incorporates a suitable loss term. Variant (iv) that replaces the hypernetwork $f$ with a ResNet/Transformer $f'$ is less competitive in all metrics, except the IOU score, where it outperforms the baselines but not the other variants of our method.

Comparing the two alternative architectures of $f$, the Transformer design outperforms the ResNet design in almost all cases. A notable exception is when hypernets are not used. However, in this case the overall performance is low.

Fig.~\ref{fig:sample} presents sample results for our method. As can be seen, in the final solution, the metallic region constraints are respected, and the final radiation pattern is more similar to the requirement than the intermediate one obtained from the hypernetwork before the refinement stage.

{\bf Antenna Arrays\quad} For the Antenna Arrays experiments, the synthetic dataset used for the single antenna case was repurposed by sampling multiple antennas. For each instance, we selected 
(i) the number of elements in the array, uniformly between 1 and 6, (ii) single antennas from the synthetic dataset, and (iii) the position of each element. In order to match the real-world design, we made sure no antenna is selected more than once (the probability of such an event is admittedly small). The array gain was computed based on Eq.~\ref{eq:AG}. All the ablations and our method were trained only over the train set of the synthetic dataset. For testing, we employed a similarly constructed syntehtic dataset, as well as two different fabricated antennas \cite{chen_design_2018,singh_design_2016}. In addition, in order to ensure that our suggestion solves a real-world problem, we evaluate the network suggestion for an alternative design of iPhone 11 Pro Max's antenna array. In this case, we do not know the ground truth design. Therefore, we use a theoretic array response of isotropic elements, simulate the suggested design with openEMS \cite{openEMS}, and compare the result with the same figure of merit from the FCC report of the device\footnote{{{iPhone} 11 {Pro} {Max} {FCC} {report}}, \url{fccid.io/BCG-E3175A}}.

 We apply our method with both architectures of $f$, and with $Q=10,000$ or when $q$ determines all of the parameters of $f$ ($Q=\inf$). In addition to the nearest neighbor baseline, which performs retrieval from the training set by searching for the closest example in terms of highest $msSSIM$ metric  of the input's $AG_{input}$ and the sample's $AG_{nn}$ . We also consider the following baselines: (i) a baseline without a hyperhypernetwork, consisting of $f$ and $g$. (ii) A no hypernetwork variant that combines $f$ and $g$ to a single network, by adding a linear layer to arrange the dimensions of the embedding before the MLP classifier.

The results on the synthetic dataset 
are reported in Tab.~\ref{tab:results_array}. As can be seen, the full method outperforms the baseline and ablation methods. In addition, the Transformer based architectures outperforms the ResNet variants.  The additional gain in performance when predicting all of $\theta_f$ ($Q=\inf$), if exists, is relatively small. We note that this increases the training time from 2 (3) hours to  7 (10) hours for the ResNet (Transformer) model.  
 
Fig.~\ref{fig:initial}(a) presents the training loss as a function of epoch for the hyperhypernetwork that employs the Transformer hypernet ($Q=10^4$), with the different initialization techniques. See appendix for the ResNet case and further details. The hyperhypernetwork fan-in method shows better convergence and a smaller loss than both hypernetwork-fan-in \cite{Chang2020Principled} and Xavier. 

Fig.~\ref{fig:initial}(b) presents the score $H_j$ that is being used to select parameters that are predicted by $q$. Evidently, there is a considerable amount of variability between the layers in both network architectures.


Fig.~\ref{fig:array_sample}(a,b) presents sample results for our method. The metallic structure probability $O_2$ is shown in (a) in log scale, and the constraint plane $C$ (Eq.~\ref{eq:constraint}) is marked in black. As required, the probabilities are very small in the marked regions. Panel (b) presents the hard decision based on $O_1$.  Misclassified points (marked in red) are relatively rare.

Tab.~\ref{tab:results_fab} presents the reconstruction of two real-world antennas: a slotted patch antenna \cite{chen_design_2018}, and a generic patch antenna \cite{singh_design_2016}. The results clearly show the advantage of our method upon the rest of the baselines and ablations in reconstructing correctly the inner structure of these examples, while preserving the constraint of localization of the array elements.  Fig.~\ref{fig:array_sample}(c,d) show our method results for reconstructing real fabricated slotted patch antenna. See appendix for the ablation results; our results are much more similar to the ground truth design than those of the ablations.

Tab.~\ref{tab:results_iphone} presents our method's result, designing an antenna array that complies  with the iPhone 11 Pro Max physical constraints. The resulting array was simulated and compared with the reported directivity (max of Eq.~\ref{eq:D} over all directions) in Apple's certificate report. Our method achieved very high scores on both directivity and compliance to the physical assembly constraints.

\section{Conclusions}

We address the challenging tasks of designing antennas and antenna arrays, under structural constraints and radiation requirements. These are known to be challenging tasks, and the current literature provides very limited solutions. Our method employs a simulation network that enables a semantic loss in the radiation domain and a hypernetwork. For the design of antenna arrays, we introduce the hyperhypernetwork concept and show how to initialize it and how to select to which weights of the inner hypernetwork it applies. Our results, on both simulated and real data samples, show the ability to perform the required design, as well as the advantage obtained by the novel methods.


\section*{Acknowledgments}
This project has received funding from the European Research Council (ERC) under the European Unions Horizon 2020 research and innovation programme (grant ERC CoG 725974).

\bibliography{antenna}

\begin{thebibliography}{29}
\providecommand{\natexlab}[1]{#1}
\providecommand{\url}[1]{\texttt{#1}}
\expandafter\ifx\csname urlstyle\endcsname\relax
  \providecommand{\doi}[1]{doi: #1}\else
  \providecommand{\doi}{doi: \begingroup \urlstyle{rm}\Url}\fi

\bibitem[Anguera et~al.(2013)Anguera, Andújar, Huynh, Orlenius, Picher, and
  Puente]{anguera_advances_2013}
Anguera, J., Andújar, A., Huynh, M.-C., Orlenius, C., Picher, C., and Puente,
  C.
\newblock Advances in {Antenna} {Technology} for {Wireless} {Handheld}
  {Devices}, 2013.

\bibitem[Bertinetto et~al.(2016)Bertinetto, Henriques, Valmadre, Torr, and
  Vedaldi]{bertinetto2016learning}
Bertinetto, L., Henriques, J.~F., Valmadre, J., Torr, P., and Vedaldi, A.
\newblock Learning feed-forward one-shot learners.
\newblock In \emph{Advances in Neural Information Processing Systems}, pp.\
  523--531, 2016.

\bibitem[{Bogale} \& {Le}(2016){Bogale} and {Le}]{Bogale_massive_mimo}
{Bogale}, T.~E. and {Le}, L.~B.
\newblock Massive mimo and mmwave for 5g wireless hetnet: Potential benefits
  and challenges.
\newblock \emph{IEEE Vehicular Technology Magazine}, 11\penalty0 (1):\penalty0
  64--75, 2016.

\bibitem[Bulus(2014)]{bulus2014center}
Bulus, U.
\newblock Center wavelength design frequency [antenna designer's notebook].
\newblock \emph{IEEE Antennas and Propagation Magazine}, 56\penalty0
  (5):\penalty0 167--169, 2014.

\bibitem[Chang et~al.(2020)Chang, Flokas, and Lipson]{Chang2020Principled}
Chang, O., Flokas, L., and Lipson, H.
\newblock Principled weight initialization for hypernetworks.
\newblock In \emph{Int. Conf. on Learning Representations}, 2020.

\bibitem[Chen \& Lin(2018)Chen and Lin]{chen_design_2018}
Chen, W. and Lin, Y.
\newblock Design of \$2{\textbackslash}times 2\$ {Microstrip} {Patch} {Array}
  {Antenna} for {5G} {C}-{Band} {Access} {Point} {Applications}.
\newblock In \emph{2018 {IEEE} {International} {Workshop} on
  {Electromagnetics}:{Applications} and {Student} {Innovation} {Competition}
  ({iWEM})}, pp.\  1--2, August 2018.

\bibitem[Clevert et~al.(2015)Clevert, Unterthiner, and
  Hochreiter]{clevert2015fast}
Clevert, D.-A., Unterthiner, T., and Hochreiter, S.
\newblock Fast and accurate deep network learning by exponential linear units.
\newblock \emph{arXiv preprint arXiv:1511.07289}, 2015.

\bibitem[Dantzig(1955)]{dantzig_1955}
Dantzig, T.
\newblock \emph{Number: The language of science}.
\newblock Penguin, 1955.

\bibitem[{Federal Communications Commission}(2015)]{fcc_sar}
{Federal Communications Commission}.
\newblock Specific absorption rate (sar) for cellular telephones.
\newblock \emph{Printed From Internet May}, 20:\penalty0 2, 2015.

\bibitem[Ha et~al.(2016)Ha, Dai, and Le]{ha2016hypernetworks}
Ha, D., Dai, A., and Le, Q.~V.
\newblock Hypernetworks.
\newblock \emph{arXiv preprint arXiv:1609.09106}, 2016.

\bibitem[Hornby et~al.(2006)Hornby, Globus, Linden, and Lohn]{Hornby2006NASA}
Hornby, G., Globus, A., Linden, D., and Lohn, J.
\newblock Automated antenna design with evolutionary algorithms.
\newblock In \emph{Space 2006}, pp.\  7242. {NASA}, 2006.

\bibitem[Jayakumar et~al.(2020)]{jayakumar2020multiplicative}
Jayakumar, S.~M. et~al.
\newblock Multiplicative interactions and where to find them.
\newblock In \emph{International Conference on Learning Representations}, 2020.

\bibitem[Kendall et~al.(2018)Kendall, Gal, and Cipolla]{kendall2018multitask}
Kendall, A., Gal, Y., and Cipolla, R.
\newblock Multi-task learning using uncertainty to weigh losses for scene
  geometry and semantics.
\newblock In \emph{IEEE conf. on computer vision and pattern recognition}, pp.\
   7482--7491, 2018.

\bibitem[Kingma \& Ba(2014)Kingma and Ba]{kingma2014adam}
Kingma, D.~P. and Ba, J.
\newblock Adam: A method for stochastic optimization.
\newblock \emph{arXiv preprint arXiv:1412.6980}, 2014.

\bibitem[Liebig(2010)]{openEMS}
Liebig, T.
\newblock {openEMS} - open electromagnetic field solver, 2010.
\newblock URL \url{https://www.openEMS.de}.

\bibitem[Littwin \& Wolf(2019)Littwin and Wolf]{Littwin_2019_ICCV}
Littwin, G. and Wolf, L.
\newblock Deep meta functionals for shape representation.
\newblock In \emph{The IEEE International Conference on Computer Vision
  (ICCV)}, October 2019.

\bibitem[{Liu} et~al.(2014){Liu}, {Aliakbarian}, {Ma}, {Vandenbosch}, {Gielen},
  and {Excell}]{geneticalgo}
{Liu}, B., {Aliakbarian}, H., {Ma}, Z., {Vandenbosch}, G. A.~E., {Gielen}, G.,
  and {Excell}, P.
\newblock An efficient method for antenna design optimization based on
  evolutionary computation and machine learning techniques.
\newblock \emph{IEEE Transactions on Antennas and Propagation}, 62\penalty0
  (1):\penalty0 7--18, 2014.

\bibitem[Miron \& Miron(2014)Miron and Miron]{miron_small_2014}
Miron, D.~B. and Miron, D.~B.
\newblock \emph{Small {Antenna} {Design}.}
\newblock Elsevier Science, Saint Louis, 2014.
\newblock ISBN 9780080498140.

\bibitem[Misilmani \& Naous(2019)Misilmani and Naous]{misilmani_machine_2019}
Misilmani, H. M.~E. and Naous, T.
\newblock Machine {Learning} in {Antenna} {Design}: {An} {Overview} on
  {Machine} {Learning} {Concept} and {Algorithms}.
\newblock In \emph{2019 {International} {Conference} on {High} {Performance}
  {Computing} \& {Simulation} ({HPCS})}. IEEE, July 2019.

\bibitem[Parmar et~al.(2018)Parmar, Vaswani, Uszkoreit, Kaiser, Shazeer, Ku,
  and Tran]{parmar_image_2018}
Parmar, N., Vaswani, A., Uszkoreit, J., Kaiser, L., Shazeer, N., Ku, A., and
  Tran, D.
\newblock Image transformer.
\newblock In \emph{International Conference on Machine Learning}, 2018.

\bibitem[Poor \& Wornell(1998)Poor and Wornell]{poor_wireless_1998}
Poor, H.~V. and Wornell, G.~W. (eds.).
\newblock \emph{Wireless communications: signal processing perspectives}.
\newblock Prentice {Hall} signal processing series. Prentice Hall PTR, Upper
  Saddle River, N.J, 1998.
\newblock ISBN 9780136203452.

\bibitem[{Prado} et~al.(2018){Prado}, {López-Fernández}, {Arrebola}, and
  {Goussetis}]{svm_antenna}
{Prado}, D.~R., {López-Fernández}, J.~A., {Arrebola}, M., and {Goussetis}, G.
\newblock Efficient shaped-beam reflectarray design using machine learning
  techniques.
\newblock In \emph{2018 15th European Radar Conference (EuRAD)}, 2018.

\bibitem[Roberts \& Abeysinghe(1995)Roberts and
  Abeysinghe]{roberts_two-state_1995}
Roberts, J. and Abeysinghe, J.
\newblock A two-state {Rician} model for predicting indoor wireless
  communication performance.
\newblock In \emph{Proceedings {IEEE} {International} {Conference} on
  {Communications} {ICC}}, 1995.

\bibitem[Santarelli et~al.(2006)Santarelli, Yu, Goldberg, Altshuler, O'Donnell,
  Southall, and Mailloux]{Military}
Santarelli, S., Yu, T.-L., Goldberg, D., Altshuler, E., O'Donnell, T.,
  Southall, H., and Mailloux, R.
\newblock Military antenna design using simple and competent genetic
  algorithms.
\newblock \emph{Mathematical and Computer Modelling}, 43:\penalty0 990--1022,
  2006.

\bibitem[Singh(2016)]{singh_design_2016}
Singh, S.~P.
\newblock Design and {Fabrication} of {Microstrip} {Patch} {Antenna} at 2.4
  {Ghz} for {WLAN} {Application} using {HFSS}.
\newblock \emph{IOSR Journal of Electronics and Communication Engineering},
  pp.\  01--06, January 2016.

\bibitem[Vaswani et~al.(2017)Vaswani, Shazeer, Parmar, Uszkoreit, Jones, Gomez,
  Kaiser, and Polosukhin]{vaswani2017attention}
Vaswani, A., Shazeer, N., Parmar, N., Uszkoreit, J., Jones, L., Gomez, A.~N.,
  Kaiser, L., and Polosukhin, I.
\newblock Attention is all you need.
\newblock In \emph{NeurIPS}, 2017.

\bibitem[von Oswald et~al.(2020)von Oswald, Henning, Sacramento, and
  Grewe]{Oswald2020Continual}
von Oswald, J., Henning, C., Sacramento, J., and Grewe, B.~F.
\newblock Continual learning with hypernetworks.
\newblock In \emph{International Conference on Learning Representations}, 2020.

\bibitem[{Wheeler}(1975)]{H_small_antenna}
{Wheeler}, H.
\newblock Small antennas.
\newblock \emph{IEEE Transactions on Antennas and Propagation}, 23\penalty0
  (4):\penalty0 462--469, 1975.

\bibitem[Zhou~Wang \& Bovik(2003)Zhou~Wang and Bovik]{ZWang}
Zhou~Wang, E. P.~S. and Bovik, A.~C.
\newblock Multi-scale structural similarity for image quality assessment.
\newblock \emph{IEEE 37th Asilomar}, 2003.

\end{thebibliography}
\bibliographystyle{icml2021}

\appendix
\section{Experimental Results}
\begin{figure}[h]
    \includegraphics[width=0.98\linewidth]{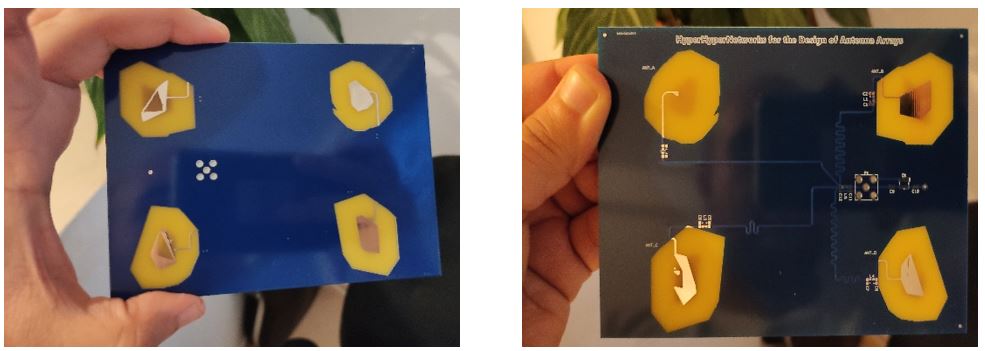}
    \vspace{-.12in}
    \caption{Manufactured antenna array. Front and back views.}
    \vspace{-.12in}
    \label{fig:array_sample_m}
\end{figure}

a sample array design was sent out for manufacturing Fig.~\ref{fig:array_sample_m}.
(i) The simulated antenna and the fabricated one agree with a 1dB difference in terms of directivity as measured in an anechoic chamber.
(ii) The measured antenna array efficiency,$\eta = 99.6\%$.
(iii) Connected to a Galaxy phone that has also debug telemetries from a cellular modem for testing, the logs extracted from the phone show that with the manufactured antenna array the SNR of receiving the cell signals improves,in comparison to the phones original antenna, by 10dB.

\section{Signal Model}
In order to select the number of antennas for the antenna array use case, we analyze a specific yet common specification of the mobile channel and its physical layer. 
For the received signal model, we assuming a QPSK transmitted narrowband symbol and Short Observation Interval Approximation (SOIA approximation), propagate through Rician channel \cite{roberts_two-state_1995}.
The justification for QPSK signal can be seen from the vast majority of protocols using it\footnote{{{Tektronix} {Overview} {of} {802.11} {Physical} {Layer}}, \url{https://download.tek.com/document/37W-29447-2_LR.pdf}}.

Let $s$ be a QPSK symbol, i.e.
 \begin{equation}
     s \in [1+j,1-j,-1+j,-1-j]/\sqrt{2}
 \end{equation} 
Let subscript $i$ denote the number of receiving antenna in the array. The physical channel $h_i$ can be modeled as a complex phasor for the i-th antenna physical channel \cite{roberts_two-state_1995}.
 \begin{equation}
     h_i \sim rice(K,\Omega)
 \end{equation}
Where $K$ is the ratio between the direct path and the reflections, $\Omega$ is the total received power.
The noise is modeled as complex normal additive noise,
 \begin{equation}
     n\sim CN(0,\sigma^2)
 \end{equation}
The received signal at the i-th antenna reads,
 \begin{equation}
     s_i = h_i\dots + n
 \end{equation}

The MIMO scheme chosen for the evaluation task is the Multi Rate Combiner (MRC) \cite{poor_wireless_1998}, 
We define the Signal to Noise Ratio (SNR) as
 \begin{equation}
     SNR_{i} = \frac{|h_{i}|^2}{|\sigma^2|}
 \end{equation}
Thus the post-processing signal $s_p$ reads
\begin{equation}
     s_p = \frac{SNR_0s_0+...SNR_is_i}{SNR_0+...SNR_i}
 \end{equation}
 
  \paragraph{Simulations}
 Let $SNR_{i} \sim 4dB, K = 2.8, Delay spread = 58 [ns]$, we ran a Monte-Carlo of Bit Error Rate (BER) as function of number of antennas in receiving. 
 \begin{figure}[]
     \centering
     \includegraphics [scale=0.5]{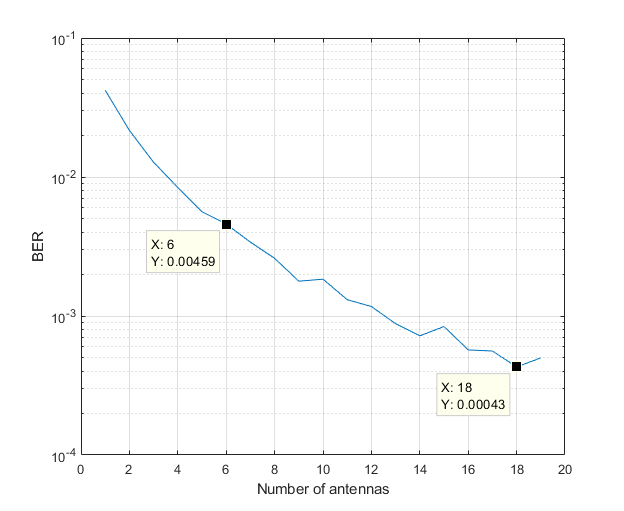}
     \caption{BER performance of MRC combiner with different number of antennas. With $10^5$ Monte Carlo iterations.}
     \label{fig:bervsnumberofantennas}
 \end{figure}
 
 From the simulation, it is observable that while increasing the number of antennas from 1 to 6 improves the Bit Error Rate (BER) by an order of magnitude, in order to achieve another order of magnitude improvement we are ought to increase the number of elements by three times. This implies that for the given (fairly common) scenario, 6 elements are a reasonable upper bound. We, therefore, set $N_{array} = 6$.
 
 The Array Gain for the observation angle $\phi$, $\theta$ reads
 \begin{equation}
     AG(\theta,\phi) = \sum_{ant}(G_{ant}(\theta,\phi) w_{ant} \exp(-jkr_{ant}))\\
     \label{eq:agang}
 \end{equation}
 \begin{equation}
     k = 2\pi/\lambda \times[sin(\theta)cos(\phi),sin(\theta)sin(\phi),cos(\theta)]
 \end{equation}
 Where $G_{ant}(\theta,\phi) \in\mathbb{R}$ is the real valued gain of an element in the array, $r_{ant}\in\mathbb{R}^3$ is the position of this element relative to zero phase of the array.
 A natural selection for $w_{ant}$ is the following beamforming coefficients
 \begin{equation}
   w_{ant} = exp(j\frac{2\pi}{\lambda}sin(\theta_d)(cos(\phi_d)r_x + sin(\phi_d)r_y))
 \end{equation}
 Where $\theta_d,\phi_d$ is the beamforming angular direction.
 Choosing the centered steered array (zero direction) Eq.~\ref{eq:agang} reads
 \begin{equation}
 \label{eq:AG2}
     AG(\theta,\phi) = \sum_{ant}(G_{ant}(\theta,\phi)\exp(-jkr_{ant}))\\
 \end{equation}
 
 \section{Reproducibility}
 \subsection{Dataset}
 Our synthetic dataset includes 3,000 examples of simulated multi-layer printed circuit board (PCB) antennas. We use \cite{openEMS} engine with the MATLAB API to generate all of those examples. In order to span a wide range of designs we opt for random designs, 
 We draw the following random parameters over uniform distribution : (i) number of polygons in each layer (ii) number of layers in the antenna (iii) number of cavities in the polygon. The feeding point is fixed during all simulations, as the dielectric constant. The dimensions of the antenna were also allowed to change in the scale of $[\frac{\lambda}{10},\frac{\lambda}{4}]$.
 
 For the array antennas a post-simulation process was done as stated in the article, where for each array training example the following parameters are chosen:
(i) a Random number of elements over uniform probability $U(1,6)$ (ii) Random position out of 6 predefined center phase locations (iii) Calculating the array gain, $AG$, according to Eq.~\ref{eq:AG2}.

 \subsection{Code}
We use the PyTorch framework, all the models are given as separate modules: 'array\_network\_transformer','array\_network\_resnet',\\'array\_network\_ablation',\\'designer\_resnet','desinger\_transformer','simulator\_network'.
The training sequence 'array\_training.py' is also given to give an example of running the models.

In our code we make a use of external repositories:
\begin{enumerate}
    \item https://github.com/FrancescoSaverioZuppichini/ResNet,
    \item https://github.com/VainF/pytorch-msssim (multi scale SSIM)
   \item https://github.com/facebookresearch/detr (for the transformer's positional encoding) 
\end{enumerate}

The code for the block selection method is 'block\_selector.py', which is a class that gets as input a PyTorch model with loss function, input, and constraints generators as input and outputs the entropy for each layer index.

\subsection{Training}
The training time of the different networks: Simulator network 10 hours, Single antenna Designer 3-7 hours (depending on the variant), Array designer 2-10 hours (depending on the variant). All the networks trained with Adam optimizer and learning rate of $10^{-4}$ with decay factor of $0.98$ every two epochs.
After training the simulator network $h$ another $10^4$ examples were generated within 5 hours. The dataset is then divided into train/test sets with a 90\%-10\% division.

\subsection{Initialization details}
The initialization scheme we purposed assumes we have beforehand the variance of the input constraints. 
In order to estimate this variance, we calculate the mean value of the constraint variance over the train set.
This mean valued variance is then used as a constant for the initialization method when constructing the network class. 

\subsection{Computing infrastructure}
The generation of the 3,000 examples takes 2 weeks over a machine with I7-9870H CPU, 40GB RAM, and Nvidia RTX graphics card. In addition, another machine was in use with 16GB RAM and Nvidia P100 graphics card.

\section{Block Selection Experiments on Hypernetworks}

In order to further evaluate our heuristic method for selecting the important blocks, we evaluate it on one of the suggested networks of \cite{Chang2020Principled}. Specifically, we test it on the "All Conv" network, trained on the CIFAR-10 dataset.

Our block selection method first predicts the important layers, based on the entropy metric, then using a knapsack and chooses the most significant layers.  

\begin{figure}[t]
\centering
  \begin{tabular}{c}
     \includegraphics[width=0.85\linewidth]{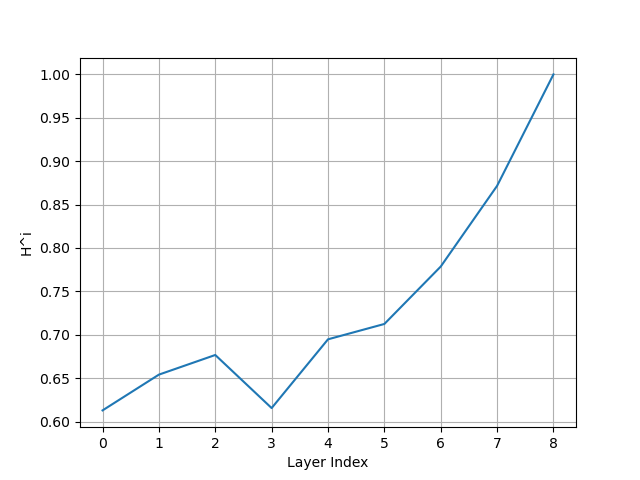} \\(a)\\
    \includegraphics[width=0.84\linewidth]{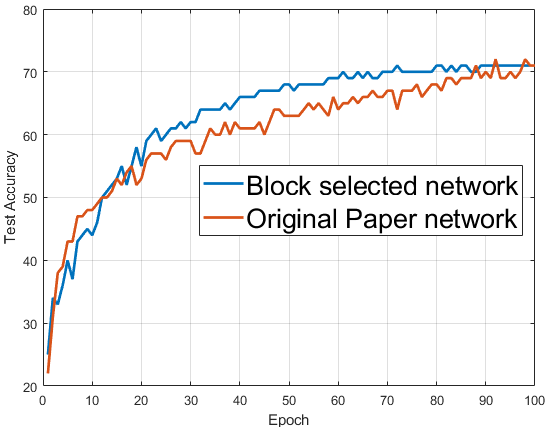}\\
     (b)\\
  \end{tabular}
    \caption{(a) Entropy loss vs layer index for the "AllConv" network. (b) Test accuracy of our block selected method (blue) and original network (orange).}
    \label{fig:blockselectionres}
\end{figure}

Fig.~\ref{fig:blockselectionres} shows (a) the entropy metric (normalized to the maximal value) for "AllConv" network, (b) Test accuracy for both networks. Using only 2.8\% of the weights (first and last layer) as the output of the hypernetwork, the suggested network is able to achieve the same results over the test set. The total size of the network is reduced by a factor of 3.4.

\section{Additional Figures}
\begin{figure*}
  \begin{tabular}{@{}c@{~}c@{}}
     \includegraphics[width=0.4830255\linewidth]{probability_of_design_certainty.png} & 
     \includegraphics[width=0.4830255\linewidth]{design_with_constraints.png} \\ 
          (a) & (b)\\
     \includegraphics[width=0.4830255\linewidth]{soltted_antenna_real.png} &
     \includegraphics[width=0.4830255\linewidth]{soltted_antenna_design.png} \\
     (c) & (d)\\
     \includegraphics[width=0.30255\linewidth]{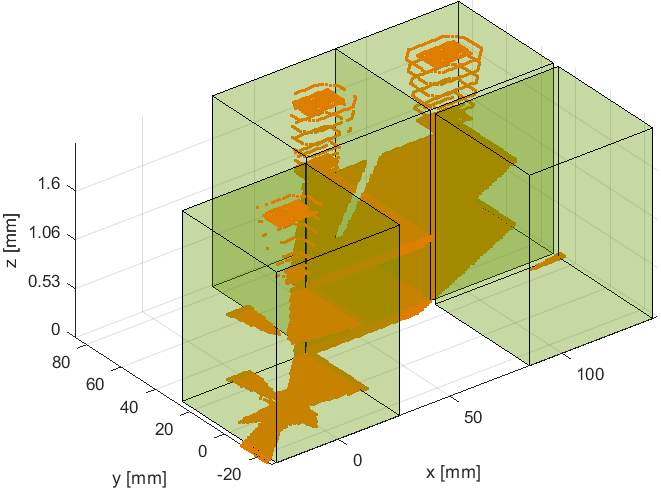}&
     \includegraphics[width=0.30255\linewidth]{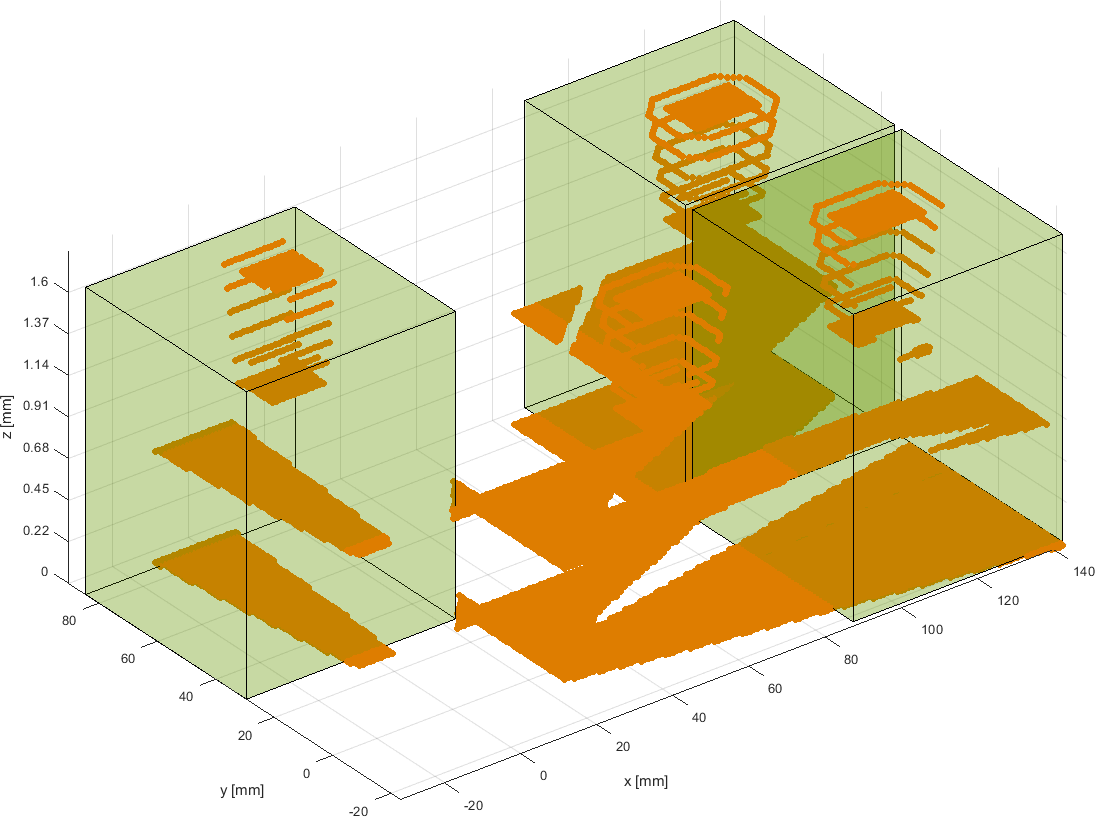}\\
     (e)&(f)\\
  \end{tabular}
    \caption{(a) The probability of points belong to a valid antenna in a synthetic test instance. The constraint plane is marked as black. (b) Same sample, the regions correctly classified as antenna are marked in brown, misclassified is marked in red.  (c) The ground truth of a slotted antenna array. (d) Our network design. (e) The design of ablation (i) that does not use a hyperhypernetwork. (f) The design of ablation (ii) that does not use a hypernetwork.}
    \label{fig:array_samplet}
\end{figure*}

Fig.~\ref{fig:array_samplet} is an enlarged version of Fig.~5 from the main paper. This figure also includes the results of the ablation experiments (panels e,f), which were omitted from the main paper for brevity.

\begin{figure}[t]
\centering
  \begin{tabular}{c}
     \includegraphics[width=0.98483\linewidth]{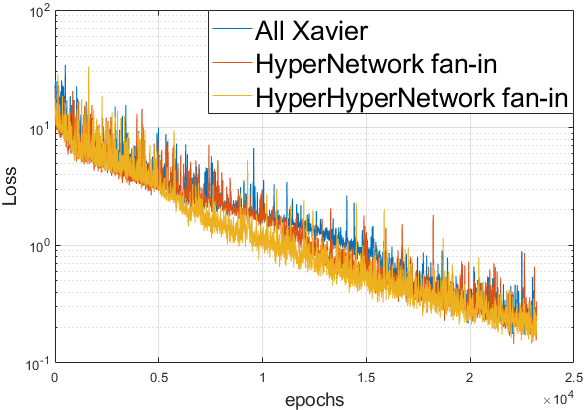} \\(a)\\
  \end{tabular}
    \caption{Loss per epochs for the different initialization scheme of $q$ (Transformer $f$).}
    \label{fig:multiplefiga}
\end{figure}

Fig.~\ref{fig:multiplefiga}  shows the effect of the different initialization schemes for the Transformer architecture (same as papers Fig.~4(a) but for a Transformer instead of a ResNet). 

\begin{figure}[t]
\centering
  \begin{tabular}{c}
    \includegraphics[width=1.1\linewidth]{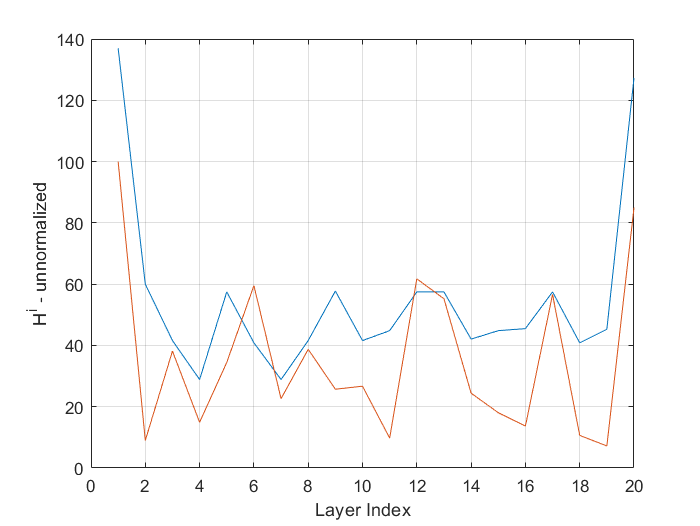}\\
  \end{tabular}
    \caption{ Un-normalized (relative to maximal value) Entropy of both ResNet and Transformer networks.}
    \label{fig:multiplefigb}
\end{figure}

Fig.~\ref{fig:multiplefigb} shows the computed entropy of different layers un-normalized, whereas in the main manuscript Fig.~4(b) depicted the same score normalized by the maximum over the different layers.

\section{Acknowledgement}
We want to thank Barak Shoafi for his assistance in manufacturing the antenna array pcb.
\end{document}